\documentclass[10pt,letterpaper]{article}
\usepackage[top=0.85in,left=2.75in,footskip=0.75in]{geometry}

\usepackage{amsmath,amssymb}

\usepackage{changepage}

\usepackage[utf8x]{inputenc}

\usepackage{textcomp,marvosym}

\usepackage{cite}
\usepackage{url}
\usepackage{nameref,hyperref}

\usepackage{microtype}
\DisableLigatures[f]{encoding = *, family = * }

\usepackage[table]{xcolor}

\usepackage{array}

\usepackage{rotating}
\usepackage{svg}

\newcolumntype{+}{!{\vrule width 2pt}}

\newlength\savedwidth

\newcommand{\comment}[2]{}

\raggedright
\usepackage[aboveskip=1pt,labelfont=bf,labelsep=period,justification=raggedright,singlelinecheck=off]{caption}

\makeatletter
\renewcommand{\@biblabel}[1]{\quad#1.}
\makeatother

\usepackage{lastpage,fancyhdr,graphicx}
\usepackage{epstopdf}
\fancyheadoffset[L]{2.25in}
\fancyfootoffset[L]{2.25in}
\begin{document}
\vspace*{0.2in}

\begin{flushleft}
{\Large
\textbf\newline{Guided-deconvolution for Correlative Light and Electron Microscopy} 
}
\newline
\\
Fengjiao Ma\textsuperscript{1,2,3},
Rainer Kaufmann\textsuperscript{4,5},
Jaroslaw Sedzicki\textsuperscript{6},
Zoltán Cseresnyés\textsuperscript{7},
Christoph Dehio\textsuperscript{6},
Stephanie Hoeppener\textsuperscript{3,8},
Marc Thilo Figge\textsuperscript{3,7,9},
Rainer Heintzmann\textsuperscript{1,2,3*}
\\
\bigskip
\textbf{1} Institute of Physical Chemistry, Abbe Center of Photonics, University of Jena, Jena, Germany
\\
\textbf{2} Leibniz Institute of Photonic Technology, Jena, Germany
\\
\textbf{3} Jena Center for Soft Matter, University of Jena, Jena, Germany
\\
\textbf{4} Centre for Structural Systems Biology, Hamburg, Germany 
\\
\textbf{5} Department of Physics, University of Hamburg, Hamburg, Germany
\\
\textbf{6} Biozentrum, university of Basel, Basel, Switzerland 
\\
\textbf{7} Applied Systems Biology, Leibniz Institute for Natural Product Research and Infection Biology - Hans Knöll Institute, Jena, Germany
\\
\textbf{8} Laboratory of Organic Chemistry and Macromolecular Chemistry, University of Jena, Jena, Germany 
\\
\textbf{9} Institute of Microbiology, Faculty of Biological Sciences, University of Jena, Jana, Germany
\\

\bigskip

%
%





* heintzmann@gmail.com

\end{flushleft}
\section*{Abstract}
Correlative light and electron microscopy is a powerful tool to study the internal structure of cells. It combines the mutual benefit of correlating light (LM) and  electron (EM) microscopy information. However, the classical approach of overlaying LM onto EM images to assign functional to structural information is hampered by the large discrepancy in structural detail visible in the LM images. This paper aims at investigating an optimized approach which we call EM-guided deconvolution. It attempts to automatically  assign fluorescence-labelled structures to details visible in the  EM image to bridge the gaps in both resolution and specificity between the two imaging modes.


\section*{Introduction}
Electron microscopy (EM) of biological samples provides the opportunity to image the structures of cells down to the level of detail of a single membrane. Yet its low specificity provides little functional information. Several technologies have been developed to specify the EM structures~\cite{bib1, bib2, bib3}. One approach is to label these structures with a fluorescent dye, and measure the same region of interest using light microscopy~\cite{bib4, bib5,bib6, bib7}, a technique called correlative microscopy. However, even with super-resolution light microscopy techniques, the light microscopy images’ resolution is still far away from the EM structural resolution.  If the correlative images are simply overlaid, it is often hard to directly identify corresponding objects and assign functional LM information to a structural EM features.

Light microscopy deconvolution~\cite{bib8,bib9,bib10,bib11,bib12,bib13} is a method that exploits the knowledge of the process of imaging, modelled as a convolution of the sample with a  point spread function (PSF) to mathematically restore the sample information. Many deconvolution methods and regularization schemes have been proposed in the past~\cite{bib14}, yet due to the ill-posed nature of the problem~\cite{bib15}, the restoration of very high spatial frequency information is still very limited.

In this manuscript, we investigate an algorithm, which we term ‘EM-guided deconvolution’, to link the LM image to its correlated EM image in a model-based approach. This method is based on the theory of maximum likelihood deconvolution. As detailed below, we observe a resolution improvement in simulations when comparing the EM-guided deconvolution results with deconvolution results based on classical priors. To validate the practical accuracy of our approach, we processed each color channel of a mixture of differently colored 40 nm beads individually using the EM-guided deconvolution approaches. Closely neighboring beads were identified with different colors, provided that image alignment was performed with great care. We then applied the EM-guided deconvolution algorithms to correlated data of biological samples. The results exhibit a realistic appearance of fluorescence-labelled membrane structures.

All the simulation and experimental data sets in this article are processed using MATLAB, with the help of the DIPimage toolbox~\cite{bib16} and Cuda~\cite{bib17} acceleration. The L-BFGS (a quasi-Newton) method of the MinFunc plugin~\cite{bib18} is used to minimise the loss function.

\section*{Principle}
The detected light microscopy image '$\mathrm{I}$' can be described as:
\begin{eqnarray}
\label{eq:Def_I}
	\mathrm{I} = f\otimes h + N,
\end{eqnarray}
where $f$ is the sample, $h$ is the point spread function (PSF). $N$ is the noise that follows the Poisson distribution, with its mean being described by the ideal image $f$ convolved with $h$.
From the theory of maximum a posterior likelihood deconvolution (MAP)~\cite{bib8}, we know that the MAP loss function is given by connecting the data term, the negative log likelihood $L(f,\mathrm{I})$ with the prior $R(f)$ through Bayes' rule, which yields for the total negative log-likelihood loss:. 
\begin{eqnarray}
\label{eq:loss}
loss = L(f,\mathrm{I}) + \lambda  R(f),
\end{eqnarray}
where $L(f,\mathrm{I})$ contains the forward- and the noise model comparing the simulated measurement with the detected image, $R(f)$ is a penalty function that accounts for the known properties of the reconstructed sample, and $\lambda$ is the coefficient which controls the strength of penalty function. The reconstructed sample is calculated by minimizing this $loss$ function. Prior knowledge about the sample being all-positive, could be implemented by a penalty term, but we chose to implement it as part of the forward model optimizing for an auxiliary function such as $f=f'^2$ mapping all numbers to the set of positive numbers, rather than optimizing for $f$~\cite{bib8}. 

The basic idea of the EM-guided deconvolution is to introduce the pre-processed and registered EM images as a position-dependent parameter in the penalty function, whereas in regular MAP deconvolution only the reconstructed image is used together with the global penalty weight $\lambda$.

\subsection*{Intensity-guided deconvolution}
In the intensity-guided deconvolution (IG) method, we directly use the intensity information of the EM image as local guidance. The EM image is pre-possessed to a binary image ($EM_{0}$) containing all the structures of interest, which could possibly correlate with a fluorescent label being present in these locations. An agreement of the deconvolution with this binary image should yield a small penalty value, whereas a disagreement should yield a larger value. We used to following penalty term:
\begin{eqnarray}
\label{eq:IG}
R(f)=\sum\frac{f_{i}}{EM\textsubscript{0}_{i} + \varepsilon},
\end{eqnarray}
where $\varepsilon$ is a small value adjusting the contrast of the EM guidance. This means that a small $\varepsilon$ enforces zero fluorescence intensity in non-segmented EM regions, whereas larger values are less stringent on enforcing darkness.
 
This ad hoc definition is a simple way to introduce the EM brightness information into the deconvolution algorithm. However, we observed an overly strong dependence on the parameter $\varepsilon$. By introducing the entropy distribution:
\begin{eqnarray}
\label{eq:EG}
R(f)=\sum f\ln\left(\frac{f_{i}}{\mathrm{e}(EM\textsubscript{0}_{i} + \varepsilon)},\right)
\end{eqnarray}
we obtained a less critical dependence. Here $\varepsilon$ is a small value adjusting the weight of the EM guidance and $\mathrm{e}$ is Euler's constant. We term this approach the entropy-guided deconvolution (EG).

\subsection*{Gradient-guided deconvolution (GG)}
A less direct way is to exploit the boundary information of objects in the EM images. That boundary information can be quantified by calculating the spatial gradient of the EM images based on the pixels.
$EM_{1} = \nabla_{xyz} EM $ where $EM$ is the pre-processed EM image. Then, the gradient image is normalized to $[0,1]$:
\[EM_{G} = \frac{EM_{1}-min(EM_{1})}{max(EM_{1})-min(EM_{1})}, \]
With the new image $EM_{G}$ we define the penalty function of the method gradient-guided deconvolution (GG) as:
\begin{eqnarray}
\label{eq:GG}
R(f)=\sum \frac{|{\nabla f}|^2_{i}}{EM\textsubscript{G}^n_{i} + \varepsilon},
\end{eqnarray}
where the small value $\varepsilon$ is controlling the strength of the EM guidance. The power $n$ is used to balance the uneven strength of the guidance inside the guidance image. We set $n=2$ as default, thus it has the same form as the numerator.

\section*{Simulation}
\begin{figure}[ht]
    \includegraphics[width=\textwidth]{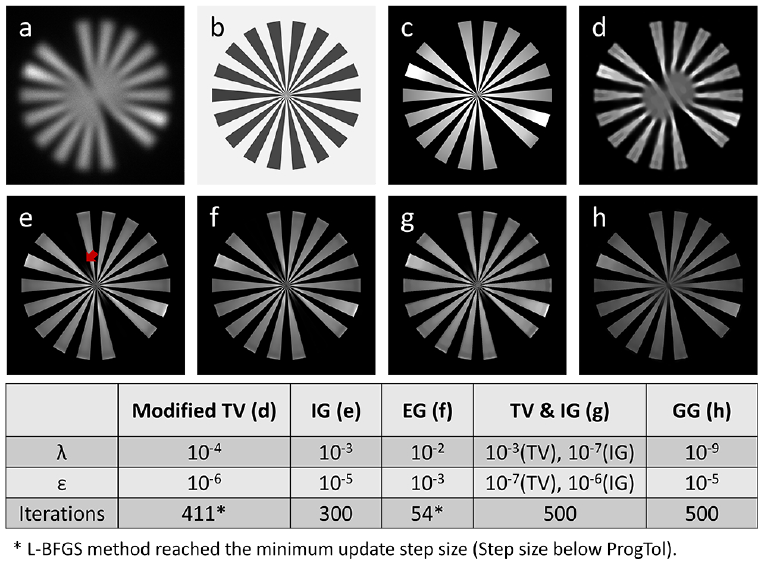}
    \caption{{\bf Deconvolution with the simulated example.}
A comparison of the results of different deconvolution methods. a) The simulated EM image correlates with the LM image. b) The simulated LM image with Poisson noise correlates with the EM image.  c) The ground truth of the LM image. d) - h) Results of modified total variation, intensity-guided, entropy-based intensity-guided, combined total variation with intensity-guided and gradient-guided deconvolution.}
    \label{simu}
\end{figure}

Here we use a Siemens star as ground truth sample to simulate the EM image in Fig.~\ref{simu}a. For the corresponding LM version of the sample (Fig.~\ref{simu}b) we used the same Siemens star but removed two spokes (Fig.~\ref{simu}c) and introduced for smooth variations in emission intensity over each spoke. The maximum expected number of photons per pixel is 1000. To avoid a problem caused by the high-frequency noise, we started our iterative deconvolution with a uniform sample estimate, equal to the mean value of the LM image.

\begin{figure}[ht]
    \includegraphics[width=\textwidth]{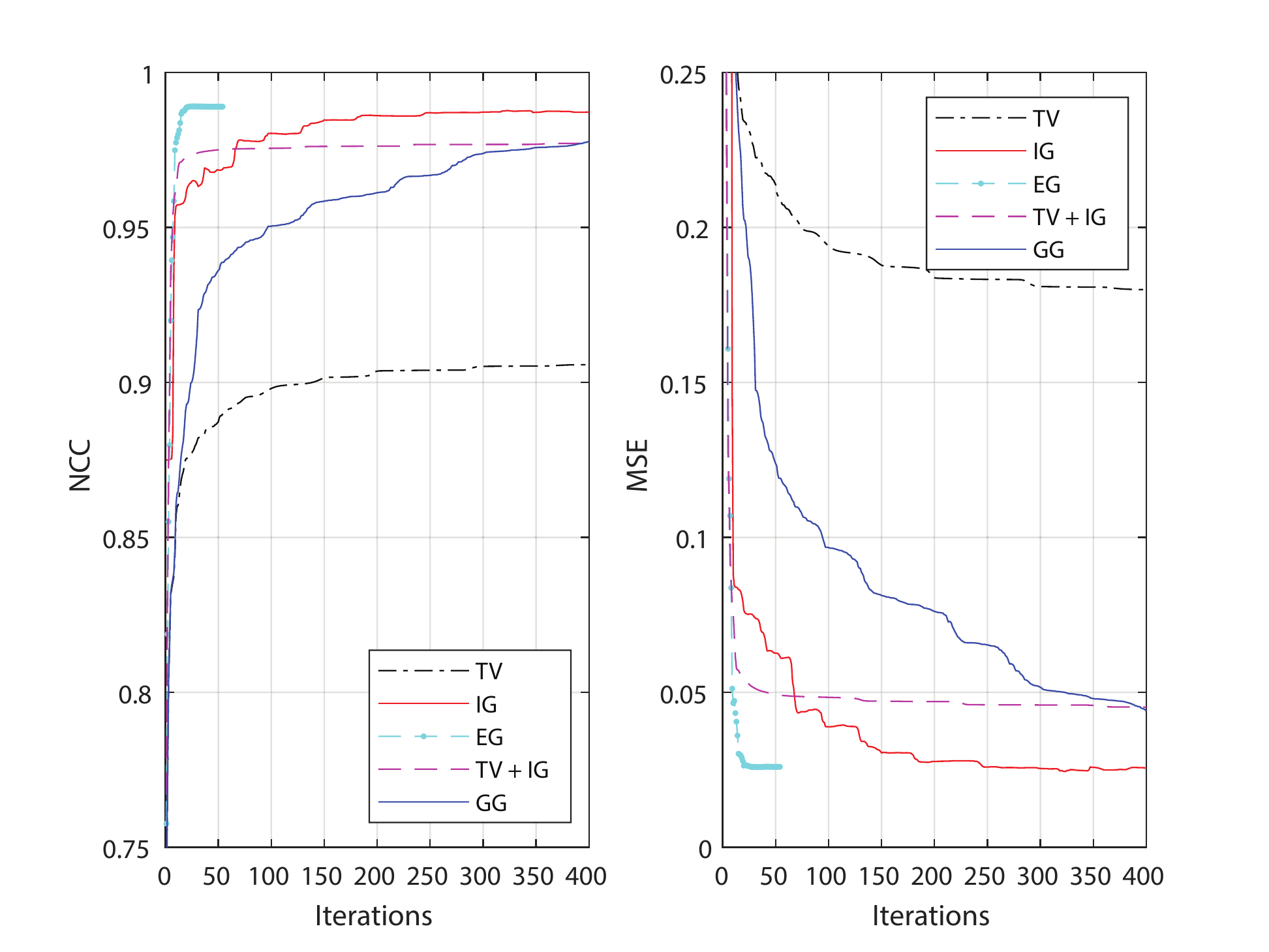}
    \caption{{\bf Error analysis of deconvolution methods.}
The restored image is compared to the ground truth after each iteration. Shown on the left is the plot of the normalized cross-correlation. The right graph is the plot of the mean square error which is normalized to the value of comparing the ground truth to the initial image. A larger (smaller) value in the normalized cross-correlation (mean square error) means a higher similarity. The parameters we select are the same as shown in the table in Fig.~\ref{simu}. With these parameters, the algorithms generate the best restorations in visualization.}
    \label{err}
\end{figure}

We observed a significant improvement in the quality of the EM-guided deconvolution results (Figs.~\ref{simu}e - ~\ref{simu}h), compared to regular deconvolution (Fig.~\ref{simu}d). Here, we use the modified TV deconvolution~\cite{bib19} as an example for the regular deconvolution. For more details on the dependence of the results on adjustable parameters, see (Fig.~\nameref{IG} - Fig.~\nameref{GG}). The corresponding parameters are listed in the table below these figures. With the guidance, the borders of the structures become clear for those large structures. The EM-guided deconvolution can also restore the small structures which are well represented by conventional deconvolution. From the plot of the normalized cross-correlation (the mean square error) comparing the restored image to the ground truth (Fig.~\ref{err}), we see that classical approaches quickly reach a steady state, whereas the EM-guided methods continues to increase (decrease), improving similarity. The EG method stopped at slightly more than 50 iterations, since the L-BFGS algorithm reached its smallest possible step size (Step Size below ProgTol). The small decrease (increase) of IG in normalized cross correlation (NCC) after hundreds of iterations indicates slightly too little regularization (over-fitting).   

The result of IG deconvolution is closely related to the selected range of parameters. If $\lambda$ is much larger than $\varepsilon $, the deconvolution provides a solution that over emphasizes the EM guidance (see~\nameref{IG}a). 
If $ \lambda$ is slightly larger than $\varepsilon$, the result shows a very good description of the intensity distribution (Fig.~\ref{simu}e). 
If $\lambda$ approximates $\varepsilon$, the algorithm will quickly run into the over-deconvolution problem, leading to the absence of high-frequency information (see the center of ~\nameref{IG}b). Thus, the values of NCC become smaller for the final result, if run for a large number of iterations.
If $\lambda$ is smaller than $\varepsilon$, the algorithm cannot restore internal brightness variations very well (~\nameref{IG}c).
If $\lambda$ is much smaller than $\varepsilon$, the effects of the penalty fades and it yields a restoration similar to that without any constraint. 

The result using IG regularization is very sensitive to the precise choice of $\lambda_{IG}$ and $\varepsilon$. To obtain less sensitive results, we recommend using is together with another classical penalty term. For instance, adding a small constraint of IG regularization to the TV deconvolution.
\begin{eqnarray}
\label{eq:TVIG}
\lambda R(f)=\lambda_{IG}R_{IG}(f)+\lambda_{TV}R_{TV}(f).
\end{eqnarray}
This yields a merge of TV and IG penalties, enforcing constant areas, yet being efficiently guided by small structural detail. This combination forces the algorithm to propagate the flatness along the non-labelled spoke, removing it also well at locations of dense structural detail (compare the location indicated by the red arrow in Fig.~\ref{simu}e and the corresponding part in Fig.~\ref{simu}g and~\nameref{IT}b).

The EG deconvolution is, compared to IG, less sensitive to the parameter settings. The algorithm reaches a similar result because of the same underlying EM guidance data and approach. Yet, due to the logarithmic term in the penalty function, the parameters in the EG method do not have such a dramatic influence on the final restored image. The value of $\lambda$ will mostly influence the speed of convergence, if roughly in the right range (~\nameref{EG}).

The gradient-guided deconvolution has more freedom to which regions the fluorescence is assigned to, since only spatial boundaries but no preferred assignments to specific segmented EM-structures are enforced. If the boundary information of the object is accurately provided, the restored image can recover the morphological characteristics of the fluorescence labeled structures surprisingly well. To obtaining a good reconstruction, the algorithm requires $\lambda$ to be smaller than $\varepsilon$. It takes more iterations than the other EM-guided methods to reach convergence.

\section*{Experimental results}
\subsection*{Beads sample}
\begin{figure}[ht]
    \includegraphics[width=\textwidth]{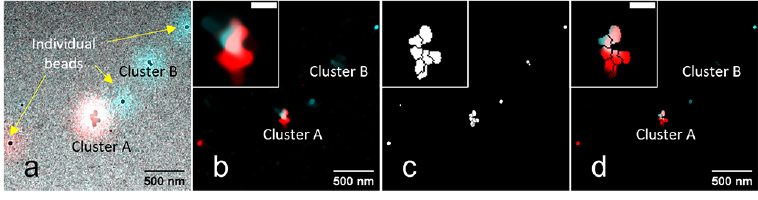}
    \caption{{\bf Overlay of CLEM images of fluorescent beads and its total variation deconvolution.} a) The identification of each beads' colour is difficult in the overlay of the CLEM image. b) The result of modified total variation deconvolution after 100 iterations, at $ \lambda = 10^{-7}, \varepsilon = 10^{-10}$. c) The binary mask created from the EM image with watershed segmentation. It can improve the image resolution, but is not sufficient to distinguish the colours of clustered beads. d) Multiplication of (b) and (c). Morphology is visible but colours cannot clearly be identified. We clipped values above 40\% of the maximum brightness for better visualisations of small signals.}
    \label{beads}
\end{figure}

To check the performance on samples of known structure, we applied our algorithm to CLEM images of fluorescent beads. The sample was a mixture of orange (565/580 nm) and red (639/720 nm) beads. They are not identifiable based on the EM image alone because all of them have roughly the same diameter and are made of the same material. If we overlap the CLEM images, the colours of the dispersed beads can be easily determined. However, the determination becomes difficult, if the two types of beads are too close to each other, especially in clusters (Cluster A), as seen in Fig.~\ref{beads}a. We use these mulitcolour beads as a test by individually processing each color channel and comparing the results. Note that algorithms which exploit prior knowledge that multiple colours should be assigned to different EM structural detail have not been used here and remain part of future research. 

From the result of the TV deconvolution, we see an improvement in the image resolution. The individual beads are becoming clearer. However, it fails in distinguishing the beads in the clusters (Cluster B in Fig~\ref{beads}c). Even though multiplying a mask (Fig~\ref{beads}b) created from the EM image can visualize the shape of the beads, there is no improvement in assigning the LM color information (Cluster A in Fig.~\ref{beads}d). Moreover, this direct multiplication might mistakenly remove fluorescent signals if the CLEM images are not very well registered (Cluster B in Fig.~\ref{beads}d).  A measured PSF was used for the deconvolution to minimise artefacts caused by a disagreement between a theoretically calculated PSF and the ground-truth PSF.

Considering the aberration difference between LM and EM images, we performed non-rigid registration of CLEM images. The alignment of the EM image was based on matching the TV deconvolved data to EM brightness information using the software BigWarp~\cite{bib20} by adding landmark points manually on the correlative images. The beads in Cluster B are not very well registered because the TV deconvolution result could not provide sufficient information for such precise registration in this area.  

\begin{figure}[ht]
    \includegraphics[width=\textwidth]{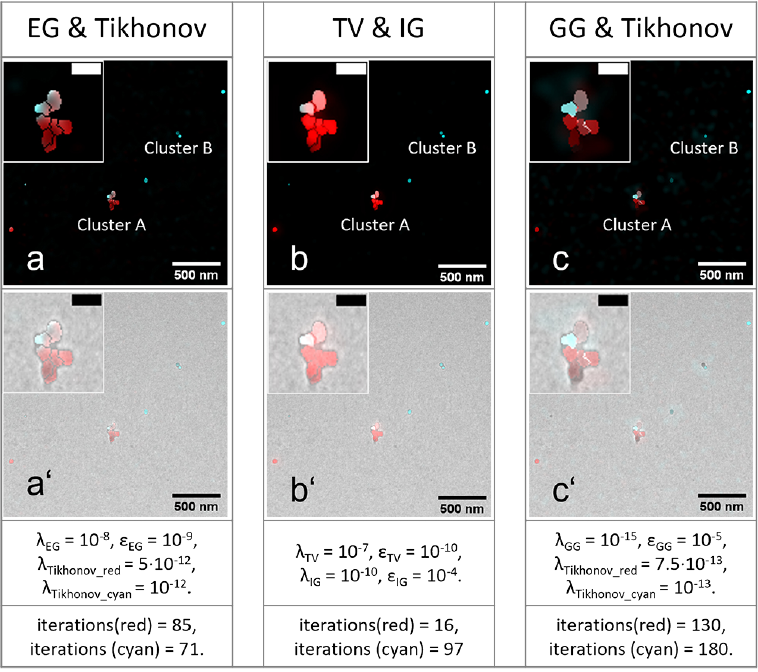}
    \caption{{\bf EM-guided deconvolution of the bead sample.}
The restorations using a) EG \& Tikhonov, b) combined TV with IG and c) GG  \& Tikhonov deconvolution. Respective overlays with the aligned EM images are presented in the panels below. The window on the top-left corner of each image shows the details of the cluster (scale bar is 100 nm). We clipped at 40\% of the maximum for a better visualization of dim structures. }
    \label{beads_res}
\end{figure}

Fig.~\ref{beads_res} shows the results of various EM-guided deconvolution methods. If the LM image is not perfectly matching to EM guidance information, the restored images can contain disturbing spike pixels. To avoid such spike pixels, we combined Tikhonov regularisation with the EG and GG methods respectively. The EG result (Fig.~\ref{beads_res}a) is quite close to the result of TV deconvolution multiplied by the binary mask (Fig.~\ref{beads}d). However, the EG deconvolution accounts for structural and functional information simultaneously. This decreases the risk accidentally removing the functional information due to a small registration error. This accidental removal of the deconvolved signal (Fig.~\ref{beads}b) can be observed in Cluster B in Fig.~\ref{beads}d after its multiplication with the misaligned mask.

The combined TV $\&$ IG method (Figs.~\ref{beads_res}b) has advantages. Although the uniformity of structure identification may be below other EM-guided methods, it requires less precise image registration. It clearly shows both beads as labelled with similar brighness (Cluster B in Fig.~\ref{beads_res}b), whereas the GG \& Tikhonov combination (Fig.~\ref{beads_res}c) shows only one of the beads.

The GG deconvolution (Figs.~\ref{beads_res}c) can more precisely separate the colour of the beads because the GG scheme allows more flexibility for the reassignment of photons during deconvolution. If the EM image is perfectly aligned, the algorithm can precisely detect that the beads at the top are cyan and the rest of the beads in Cluster A are red.  The white colour shown in the result is caused by two reasons: 
\begin{itemize}
\item The segmentation of the beads is not perfect because the beads are in contact with each other;
\end{itemize}
\begin{itemize}
\item The TEM image only shows the projection of the beads, potentially causing a 3D cluster where beads can be on top of each other looking like a single bead.
\end{itemize}
If the EM image is not well aligned, the algorithm is incapable of performing a sensible assignment. We observe only one reconstructed fluorescent bead in the deconvolution as well as a distributed cloud of fluorescence in Fig.~\ref{beads_res}c' Cluster B.

\subsection*{EM-guided deconvolution in 2D biological samples}
We applied the algorithm to correlative in-resin super-resolution fluorescence and electron microscopy imaging data~\cite{bib18}. The fluorescence microscopy image displays mVenus-labelled membrane structures (plasma membrane, endoplasmic reticulum (ER)) of a HEK293T cell embedded in resin~\cite{bib18}.

\begin{figure}[t]
    \includegraphics[width=\textwidth]{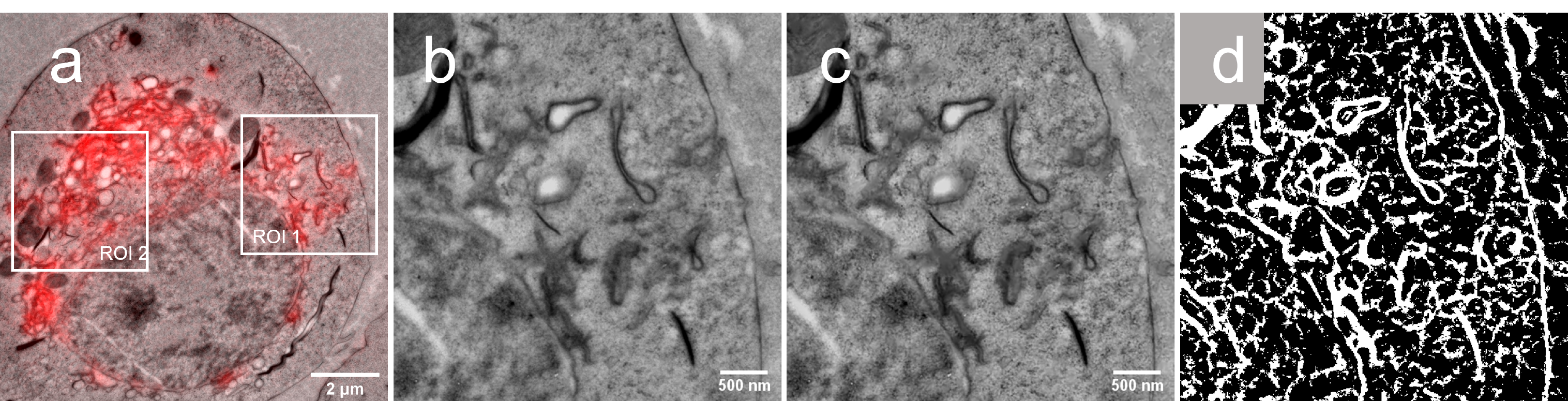}
    \caption{{\bf CLEM data of the mVenus labelled membrane structures.}
a) The overlay of the wide-field fluorescence image and the correlative TEM image remains unspecific with respect to the fluorescent labelled double-layer membranes. b) The TEM image which shows the same region of interest as (a). Membrane information (d) was extracted by machine learning from the denoised version of the TEM image (c).}
    \label{RK_CLEM}
\end{figure}

We use the wide-field image (Fig.~\ref{RK_CLEM}a), as the input image for the deconvolution. The PSF is calculated depending on the experimental parameters with the Richards $\&$ Wolf method~\cite{bib9}. The correlative LM/EM images were aligned in the ImageJ plugin BigWarp. The deconvolution algorithms are applied to the whole data sets, but we use only the cropped region marked in the white the squares (ROI1 and ROI2) to analyse the restorations. The image registration accuracy in ROI1 is higher than that in ROI2 because the membrane density in ROI1 is lower than that in ROI2, allowing us to extract more structural information from the LM image as landmarks for the CLEM image registration. To better extract the useful information from the EM image as the guidance, we used the software-trainable Weka Segmentation~\cite{bib21} on the image prepossessed by the software denoiseEM~\cite{bib22}, to deal with the complex biological structures. Figs.~\ref{RK_CLEM}b-~\ref{RK_CLEM}d use the TEM image, which is the same region as ROI1, to show the progress of the extraction of the EM image.

\begin{figure}[ht]
    \includegraphics[width=0.95\textwidth]{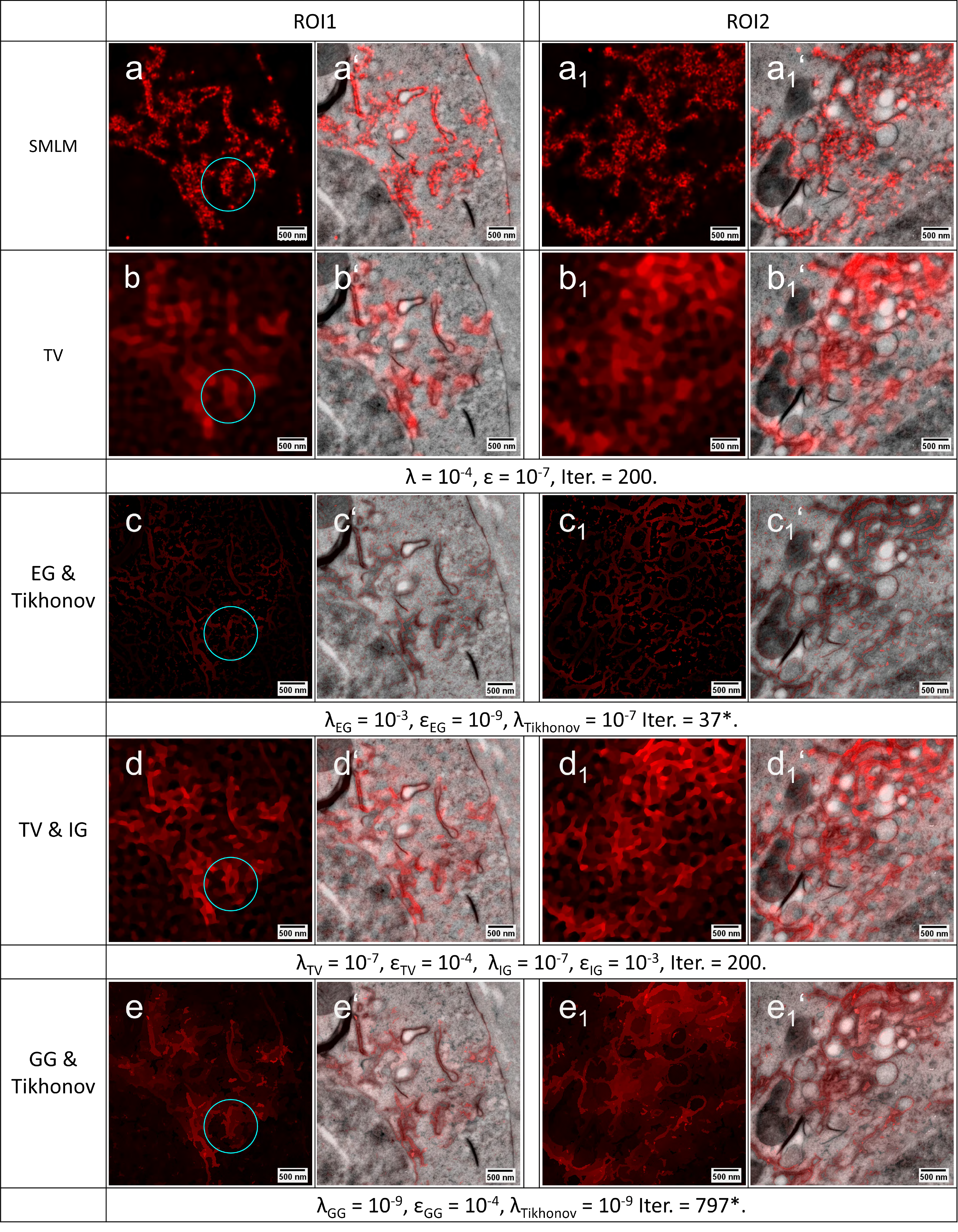}
    \caption{{\bf Restored images of ROI1 and ROI2.}
a) - a\textsubscript{1}') The restored images of single-molecule localization microscopy. b) - b\textsubscript{1}') The results of modified total variation deconvolution. c) - c\textsubscript{1}') The results of EG $\&$ Tikhonov deconvolution. b) - b\textsubscript{1}') The results of the combined TV $\&$ IG deconvolution. c) - c\textsubscript{1}') The restored images of GG $\&$ Tikhonov deconvolution.
The results of the EM-guided deconvolution shows more information than the regular LM deconvolution. Their results show good agreement with the single-molecule localization microscopy data.}
    \label{RK_result}
\end{figure}

The EM-guided deconvolution shows its robustness in the restoration of the fluorescence labelled membranes (Fig.~\ref{RK_result}c - Fig.~\ref{RK_result}e). 
The segmentation result from machine learning contains much more structural detail than needed. Such redundant information can be eliminated by EM-guided deconvolution. The EM-guided deconvolution can restore the double membranes from the wide-field image of ROI1 (the blue circles in Fig.~\ref{RK_result}). The membrane information is clearly enhanced in the GG deconvolution. It shows a very high similarity to the restored image obtained by the single-molecule localization microscopy (Fig.~\ref{RK_result}a), validating EM-guided deconvolution method. The EG deconvolution shows the benefit of restoring the structures where the membranes are dense (Fig.~\ref{RK_result}c\textsubscript{1}). In this case, the CLEM image registration is less precise due to the lack of correlative detail. In cases where other deconvolutions (Fig.~\ref{RK_result}b\textsubscript{1}, Fig. ~\ref{RK_result}c\textsubscript{1} and Fig.~\ref{RK_result}e\textsubscript{1}) create large patches of colour, the  TV \& IG deconvolution (Fig.~\ref{RK_result}d\textsubscript{1}) still convincingly assigns the fluorescence to membranes.

The digitized EM image does not necessarily need to be binarized, however a sharp edge that can describe the outer line of the objects is required. The intensity values of the EM image can be sufficient (\nameref{RK_GG}a). We can also create the guidance by labeling more than two classes (\nameref{RK_GG}c). As soon as the guidance is effective, the GG deconvolution can generate a convincing result. The restoration of the detailed structures is highly dependent on the EM guidance. Thus, there is some difference between the deconvolution results shown in \nameref{RK_GG} and Fig. ~\ref{RK_result}c.

\subsection*{EM-guided deconvolution in a biological 3D sample}
\begin{figure}[ht]
    \includegraphics[width=\textwidth]{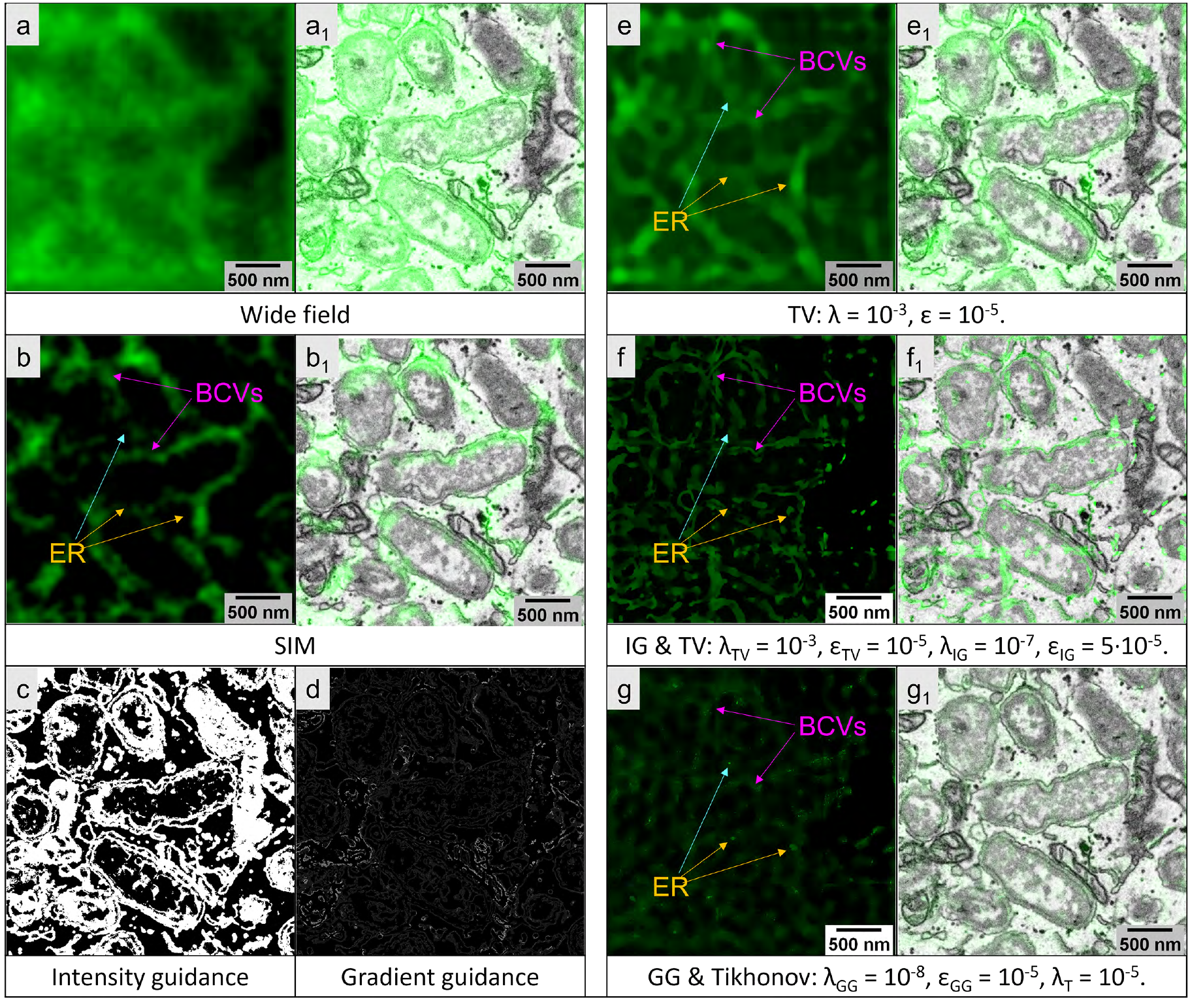}
    \caption{{\bf Cross sections of the CLEM images showing the GFP-labeled endoplasmic reticulum (ER) and \emph{Brucella}-containing vacuoles.}
    a) The wide-field fluorescence image. b) The SIM restoration image. c) The intensity guidance map is obtained via Weka segmentation. d) The gradient guidance is generated by calculating the absolute spatial gradient of the intensity guidance map. e) The restored images of the modified total variation deconvolution (brightness -20\%, contrast +40\%). f) The restoration of the combined TV $\&$ IG deconvolution from the wide-field fluorescence microscopy image (brightness +40\%, contrast -40\%). g) The restored images of the gradient-guided deconvolution (brightness +60\%). a\textsubscript{1}), b\textsubscript{1}),  e\textsubscript{1}), f\textsubscript{1}) g\textsubscript{1}): The overlay of the CLEM image.}
    \label{SEM}
\end{figure}

We then applied the EM-guided deconvolution algorithms to 3D-CLEM images of HeLa cells infected with \emph{Brucella abortus}~\cite{bib23}. The membranes of the endoplasmic reticulum (ER) as well as \emph{Brucella}-containing vacuoles (BCVs) in the host cells were labelled with the GFP-Sec61$\beta$ fusion protein. Structured illumination microscopy (SIM) was used to obtain high-resolution images of the labelled structures. The labelled structures could be identified to reside outside the bacteria (Fig. ~\ref{SEM}f, f\textsubscript{1}, g and g\textsubscript{1}). However, the resolution improvement by the SIM technology is not sufficient to clearly identify the ER markers (see Fig.~\ref{SEM}b,b\textsubscript{1}).

We used the wide-field image (Fig. ~\ref{SEM}a, a\textsubscript{1}), the sum of all the phases of the SIM dataset, as the input image for guided deconvolution, such that the SIM reconstruction can serve as validation data. The SEM image was aligned by using the eC-CLEM~\cite{bib24} based on the position of the bacteria. Since the data set was aligned rigidly and the distortion of the LM image may differ from that in the SEM image, there may be some disagreement in the details. The alignment accuracy of the center area is higher than in the rest of the image.  The data set was processed in 3D using a theoretical calculated 3D PSF. Due to the large size, the image was processed in segments, which were recombined afterwards. 

The membranes were segmented in the Weka plugin in Fiji (Fig. ~\ref{SEM}c). It can be directly used as the intensity guidance, even though the segmentation contains more information, including details of the mitochondrial membrane. The guidance for the GG deconvolution is generated by taking the gradient of the segmented image, as shown the bottom row in Fig. ~\ref{SEM}d.

The EM-guided deconvolution shows ER markers at a level of detail far beyond what can be achieved by regular deconvolution methods. Figs.~\ref{SEM}e, f, g (~\nameref{S1_Video}) show the restored images with the TV, the TV $\&$ IG and the GG deconvolution methods. From the results of the EM-guided deconvolution, we see that the GFP labelled membrane covers the bacterial cell body and some of the ER markers outside the bacteria in the host cells, which agrees well with the SIM results (see Fig. ~\ref{SEM}b). Here, we only show the result of the TV $\&$ IG deconvolution as representative of the intensity-guided deconvolution because it always provides better performance than the other intensity-related deconvolutions. The ER membranes could be restored at high quality. Compared with the IG method, the GG deconvolution is more dependent to the accuracy of the image registration. The guided deconvolution did not convincingly represent the details of the ER structures because the disagreement between the EM information and the LM information does not guide the deconvolution in the expected direction (see the blue arrows in Fig.~\ref{SEM}).



\section*{Conclusion and outlook}
In this paper, we investigate the algorithm 'EM-guided deconvolution'  for the CLEM images to automate recognizing the fluorescence information in the registered EM image. Both the intensity-guided and the gradient-guided deconvolution outperformed the state-of-the-art deconvolution of LM data alone. The intensity-guided deconvolution does not require excessively accurate image registration compared to gradient-guided deconvolution. If the images are precisely registered, the gradient-guided deconvolution yielded a better result than the intensity-guided deconvolution, yet being more susceptible to alignment errors.

There are still challenges to overcome for the EM-guided deconvolution. E.g., how to integrate super-resolution fluorescence microscopy data, such as STED, SIM or single molecule localization microscopy data, into our framework of EM-guided deconvolution is still a question because these methods often cannot be approximated by a convolution with a well-known, spatially invariant point spread function. Furthermore, there are still some other ways to define a quality metric of the reconstructed object compared to the EM data, such as using the mutual information or the structural similarity index. Especially when dealing with multiple fluorescence labels, there is still room for improvement exploiting the assignment to separate EM structures, possibly with support by deep-learning approaches. We hope that EM-guided deconvolution will be developed further and become part of the standard tool-set of correlative light and electron microscopy imaging.

\newpage
\section*{Supporting information}

\paragraph*{SI Fig. 1}
\label{IG}
\includegraphics[width=\textwidth]{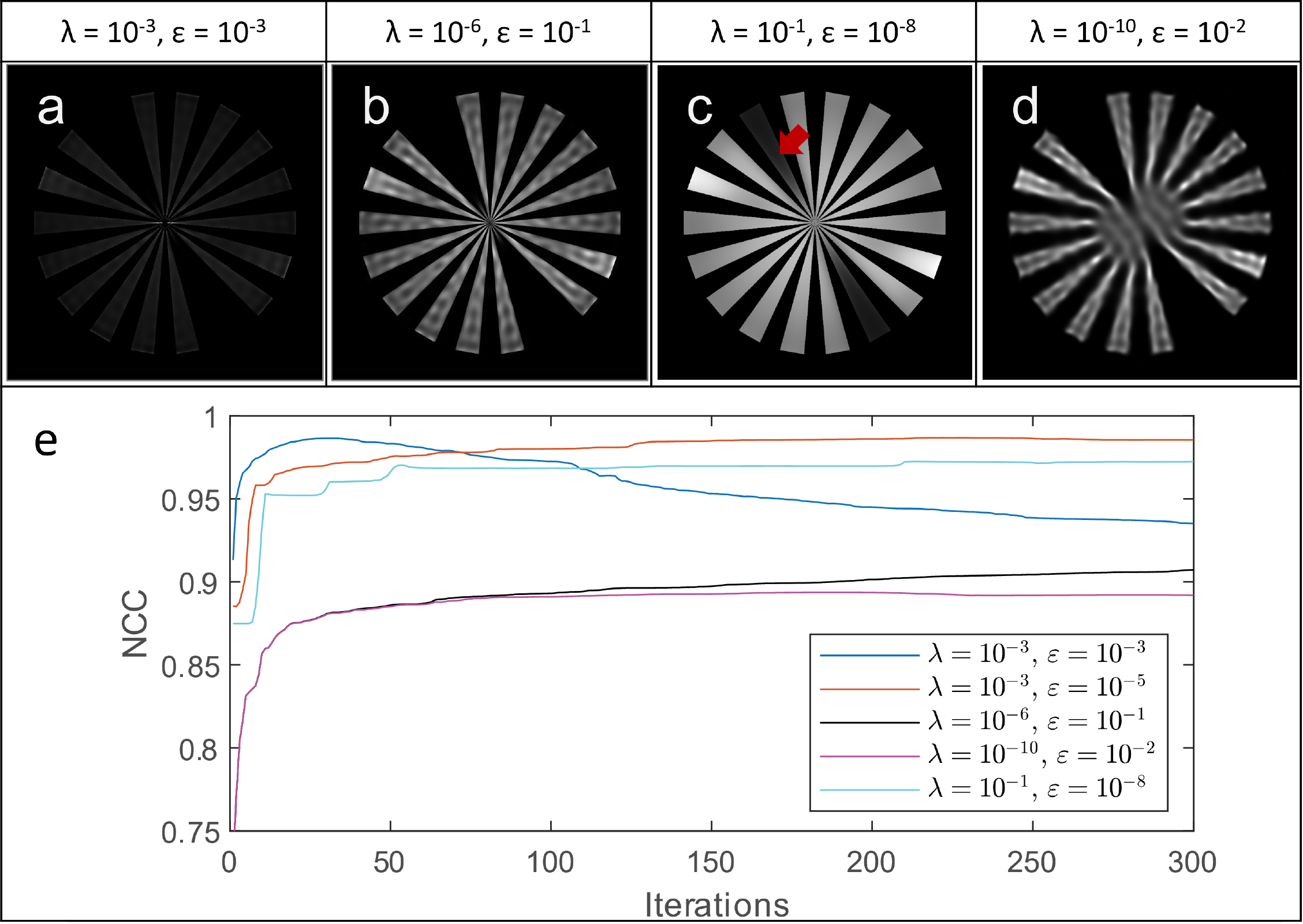}

  {\bf Restorations of IG deconvolution with different combination of parameters.} a) The restoration becomes dim if $ \lambda \simeq \varepsilon$. It yields in missing information of the structures in high-frequency. b) The algorithm will not very well restore the intensity distribution if  $\lambda < \varepsilon$.  c) The weight is too much on the EM-image which led to an incomplete removal (red arrow) of the non-fluorescent spokes if  $\lambda >> \varepsilon$. d) The EM guidance will not contribute if $\lambda$ is too small. The IG deconvolution generates good results if $\lambda > \varepsilon$ (see Figure ~\ref{simu}e). f) The NCC curves when the parameters are selected the same as in this figure.

\newpage
\paragraph*{SI Fig. 2}
\label{IT}
\includegraphics[width=\textwidth]{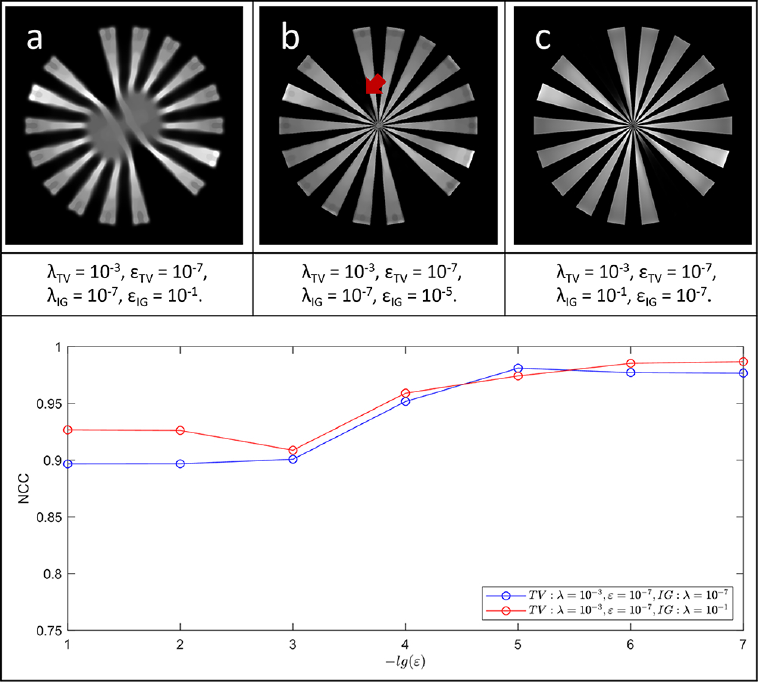}

{\bf Restorations of TV $\&$ IG method at varying strength if the IG part.} At low IG strength, a) TV regularization dominates. b) Balanced TV and IG strength yields good recovery of both, functional and morphology information. c) Strong weights of IG over emphasize the EM structural detail which overwrites the functional information.

\newpage
\paragraph*{SI Fig. 3}
\label{EG}
\includegraphics[width=\textwidth]{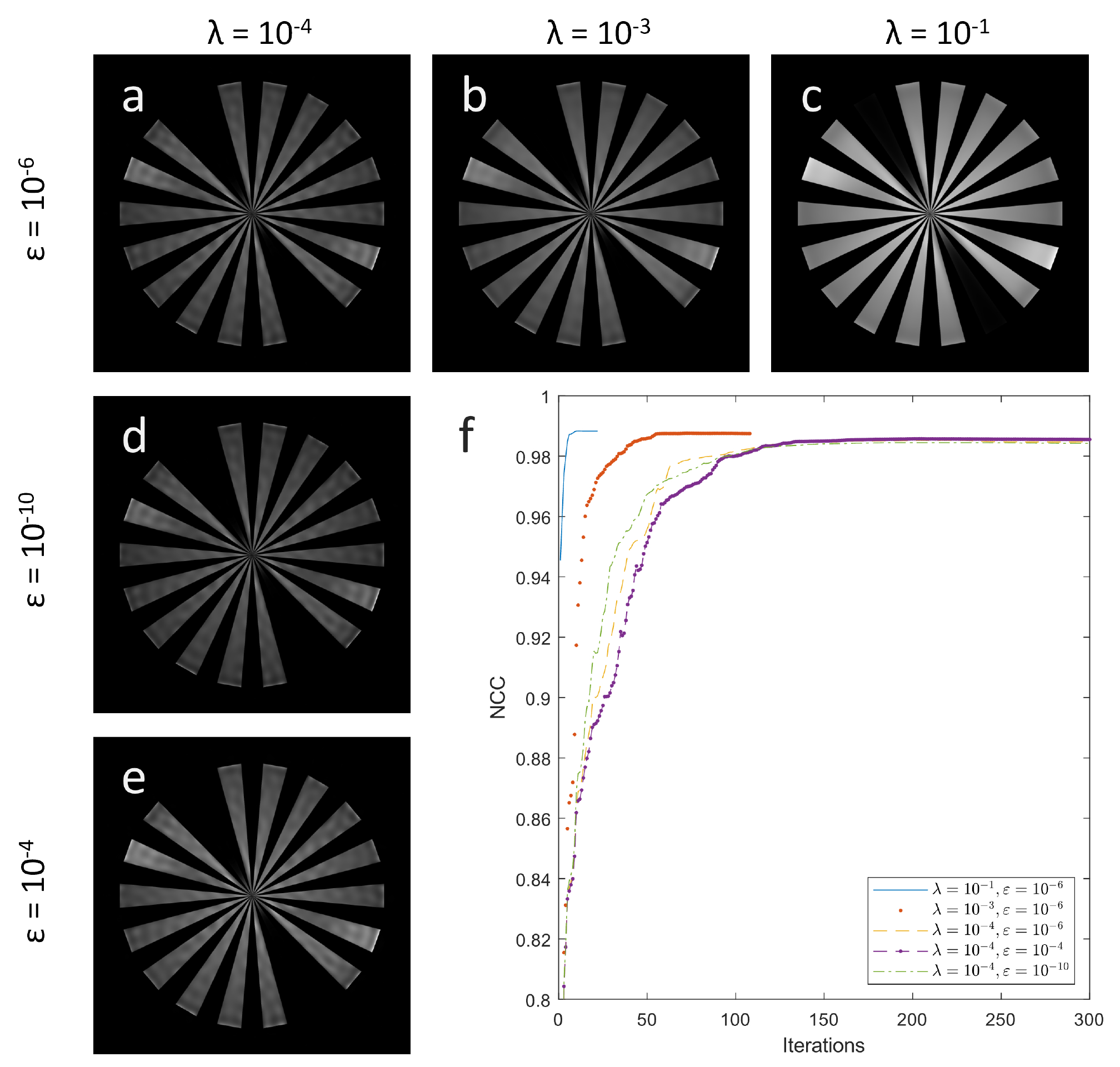}

{\bf Restorations of EG deconvolution when the parameters in different ranges.} The images in the left column are the restorations with the same $\lambda$ at various $\varepsilon$. The images in the top row are the restorations with the same $\varepsilon$ but at different $\lambda$. The final restoration of the EG deconvolution is less dependent on the parameters. Unless $\lambda$ is too large, which enforces the EM information too much, the other restored images are very similar to the ground truth in terms of NCC. A larger $\lambda$ can lead to faster convergence. The influence of $\varepsilon$ is so small that there is no perceivable difference to the final result as supported by overlapping curves.

\newpage
\paragraph*{SI Fig. 4}
\label{GG}
\includegraphics[width=\textwidth]{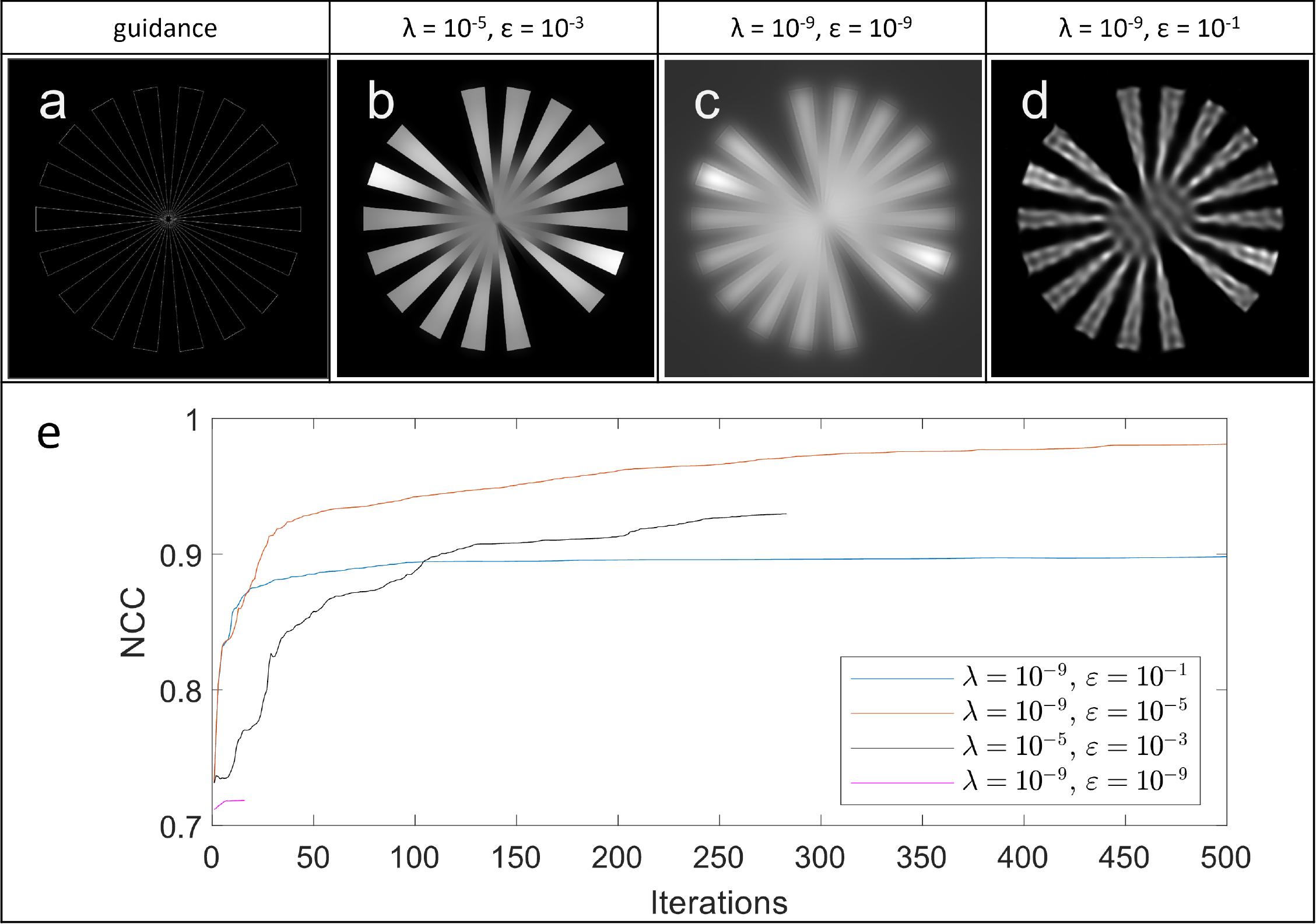}

{\bf Restorations of GG deconvolution at different parameter settings.} a) The gradient guidance. b) The restored image only reconstruct good low-frequency structures if $\lambda $ is slightly smaller than $\varepsilon$. c) The algorithm will not do restoration if $\lambda \geq \varepsilon$. d) The EM information will not provide sufficient strength for the guidance if $\lambda$ is too small. the best restoration happens when $\lambda < \varepsilon$ (see also Fig. ~\ref{simu}h).

\newpage
\paragraph*{SI Fig. 5}
\label{RK_GG}
\includegraphics[width=\textwidth]{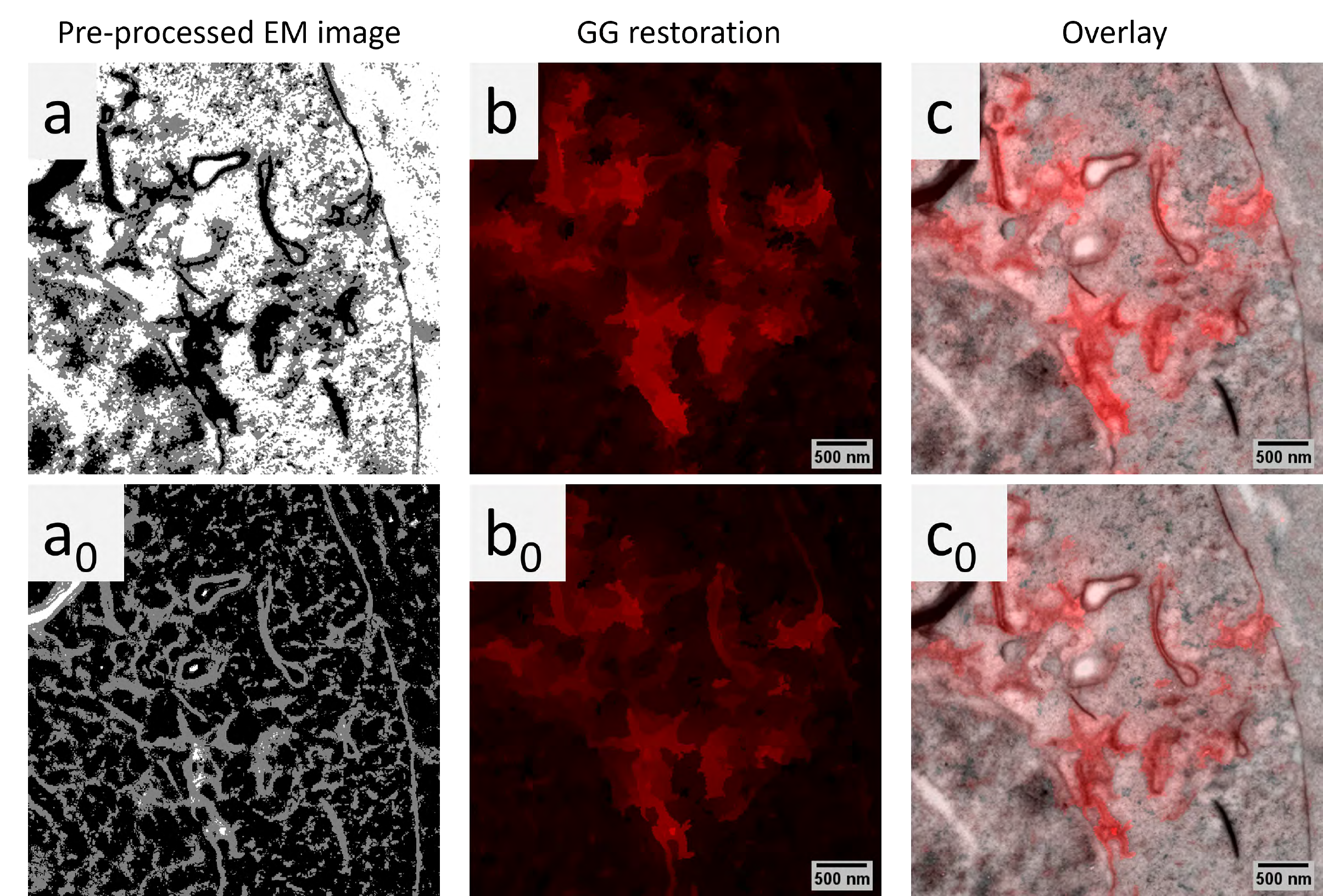}

{\bf Restored images of GG deconvolution with different EM guidance.} 
a) The pre-processed EM images. b) The GG restored images ($\lambda \textsubscript{GG}=10^{-9}$, $\varepsilon \textsubscript{GG}=10^{-4}$, $\lambda \textsubscript{Tikhonov}=10^{-8}$). c) The overlay of the restored images and the EM images. The pre-processing of the EM image was based on the Isodata algorithm~\cite{bib25}e. The result in the bottom row was generated when the EM image was trained in 3 classes with the Weka segmentation in Fiji.

\newpage
\paragraph*{SI Video.1}
\label{S1_Video}
{\bf Restorations of the bacterial sample from the stack of the wide-field images.}  The video shows the whole stack of the images corresponding to the ROI as shown in Fig.\ref{SEM}. It is a comparison of the wide-field (a), SIM restoration, total variation restoration, gradient-guided restoration and the combined total variation with intensity-guided restoration. The bottom row shows the overlay of the LM images and the EM images.  

\newpage
\paragraph*{SI Fluorescence beads sample preparation and images acquisition}

The sample solution was a mixture of orange (ThermoFisher Scientific, F8794, 565 nm /580 nm) and red (ThermoFisher Scientific, T8870, 633 nm /720 nm) fluorescence beads. The original solutions were diluted with a ratio $(1:10^{4})$ in water. The diluted solution was then mixed in a ratio of 1:1. Drop 5 µl solution on a Formvar coated TEM grid; Poly-L-Lysin was used to attach the beads to the film and ensure a sufficient distribution of the particles. After 2 minutes, the rest of the solution was removed with a paper filter (Whatman No. 1). After completely drying, the TEM grid was mounted on the cover glass with glycerin. After the light microscopy image acquisition, the TEM grid was removed from the cover glass. The TEM grid was then washed with water, such that the glycerine was removed from the TEM grid. At the end the sample was imaged by a transmission electron microscope. 

The fluorescence microscopy images were acquired by a confocal laser scanning microscope LSM980 with an $40\times$ water objective. The sample was excited by 561 nm and 639 nm lasers. The refractive index was 1.33 and the NA was 1.2. The emission wavelength of orange (red) beads is 603 nm (719 nm).  

TEM imaging was performed on a FEI Tecnai G$^{2}$ 20 (ThermoFisher Scientific) at an accelleration voltage of 200 kV. Images were recorded on a Megaview CCD camera ($1376\times1024$ image pixels, Olympus Soft Imaging Solutions, OSIS). Image intensity was adjusted to yield a max. of approximately 1000 counts.

\newpage
\section*{Acknowledgments}
This work was financially supported by the German Research Foundation (DFG) through the Collaborative Research Center PolyTarget 1278, project number 316213987, subprojects C04 and Z01. The experimental data sets of the fluorescence beads in this paper are provided from sub-project C04. This work was also supported by the Swiss National Science Foundation (SNSF, www.snf.ch) grant 310030B\_201273 to CD. We thank the Microverse Imaging Center (and Aurélie Jost / Patrick Then) for providing microscope facility support for data acquisition. The LSM 980 (used for producing the LM image in Fig.~\ref{beads} ) was funded by the Free State of Thuringia with grant number 2019 FGI 0001. The Microverse Imaging Center is funded by the Deutsche Forschungsgemeinschaft (DFG, German Research Foundation) under Germany´s Excellence Strategy – EXC 2051 – Project-ID 390713860. We also thank Martin Reifarth, Daniel Friedrich, Jennifer Lippincott-Schwartz, Harald F. Hess, Wanda Kukulski, Shen Han, Ingo Lieberwirth and Lucy Collinson for providing the experimental data sets to test the algorithms.


%
%
%

\begin{thebibliography}{10}

\bibitem{bib1}
Chklovskii DB, Vitaladevuni S, Scheffer LK. 
\newblock {Semi-automated reconstruction of neural circuits using electron microscopy}.
\newblock Current opinion in neurobiology. 2010;20(5):667–675.

\bibitem{bib2}
Lebbink MN, Geerts WJ, van der Krift TP, Bouwhuis M, Hertzberger LO, Verkleij AJ, et al. 
\newblock Template matching as a tool for annotation of tomograms of stained biological structures
\newblock Journal of Structural Biology. 2007;158(3):327–335.

\bibitem{bib3}
Xiao C, Chen X, Li W, Li L, Wang L, Xie Q, et al.
\newblock {Automatic mitochondriasegmentation for EM data using a 3D supervised convolutional network}.
\newblock Frontiers in Neuroanatomy. 2018;12:92.

\bibitem{bib4}
Ando T, Bhamidimarri SP, Brending N, Colin-York H, Collinson L, De Jonge N, et al.
\newblock {The 2018 correlative microscopy techniques roadmap}.
\newblock  Journal of physics D: Applied physics. 2018;51(44):443001

\bibitem{bib5}
Reifarth M, Preußger E, Schubert US, Heintzmann R, Hoeppener S.
\newblock {Metal–Polymer Hybrid Nanoparticles for Correlative High-Resolution Light and Electron Microscopy}.
\newblock Particle $\&$ Particle Systems Characterization. 2017;34(10):1700180

\bibitem{bib6}
Spiegelhalter C, Tosch V, Hentsch D, Koch M, Kessler P, Schwab Y, et al.
\newblock {From dynamic live cell imaging to 3D ultrastructure: novel integrated methods for high pressure freezing and correlative light-electron microscopy}.
\newblock PloS one. 2010;5(2):e9014.

\bibitem{bib7}
Ader NR, Hoffmann PC, Ganeva I, Borgeaud AC, Wang C, Youle RJ, et al.
\newblock {Molecular and topological reorganizations in mitochondrial architecture interplay during Bax-mediated steps of apoptosis}.
\newblock eLife. 2019;8:e40712.

\bibitem{bib8}
Verveer PJ, Gemkow MJ, Jovin TM. 
\newblock { comparison of image restoration approaches applied to three-dimensional confocal and wide-field fluorescence microscopy}.
\newblock Journal of microscopy. 1999;193(1):50–61.

\bibitem{bib9}
Richards B, Wolf E.
\newblock {Electromagnetic diffraction in optical systems, II. Structure of the image field in an aplanatic system}.
\newblock Proceedings of the Royal Society of London Series A Mathematical and Physical Sciences. 1959;253(1274):358–379.

\bibitem{bib10}
Tikhonov AN, Arsenin VI. 
\newblock {Solutions of ill-posed problems}.
\newblock vol. 14. Winston, Washington, DC; 1977.

\bibitem{bib11}
Gao Q, Eck S, Matthias J, Chung I, Engelhardt J, Rippe K, et al. 
\newblock {Bayesian joint super-resolution, deconvolution, and denoising of images with Poisson-Gaussian
noise}.
\newblock In: 2018 IEEE 15th International Symposium on Biomedical Imaging (ISBI 2018). IEEE; 2018. p. 938–942.

\bibitem{bib12}
Richardson WH.
\newblock {Bayesian-based iterative method of image restoration}.
\newblock JoSA. 1972;62(1):55–59.

\bibitem{bib13}
Medyukhina A, Figge MT.
\newblock {DeconvTest: Simulation framework for quantifying errors and selecting optimal parameters of image deconvolution}.
\newblock Journal of Biophotonics. 2020;13(4):e201960079.

\bibitem{bib14}
Sibarita JB. 
\newblock {Deconvolution microscopy}.
\newblock Microscopy Techniques. 2005; p. 201–243.

\bibitem{bib15}
Holmes TJ, Bhattacharyya S, Cooper JA, Hanzel D, Krishnamurthi V, Lin Wc, et al.
\newblock {Light microscopic images reconstructed by maximum likelihood deconvolution}.
\newblock In: Handbook of biological confocal microscopy. Springer; 1995. p. 389–402.

\bibitem{bib16}
Hendriks CL, Van Vliet L. D
\newblock DIPimage user manual: a scientific image processing toolbox; 2001.

\bibitem{bib17}
Heintzmann R. 
\newblock {CudaMat; 2009. https://github.com/RainerHeintzmann/CudaMat}.

\bibitem{bib18}
Johnson E, Seiradake E, Jones EY, Davis I, Gr{\"u}unewald K, Kaufmann R.
\newblock {Correlative in-resin super-resolution and electron microscopy using standard fluorescent proteins}.
\newblock Scientific reports. 2015;5(1):1–9.

\bibitem{bib19}
Soulez F, Denis L, Tourneur Y, Thi{\'e}baut {\'E}.
\newblock {Blind deconvolution of 3D data in wide field fluorescence microscop}.
\newblock . In: 2012 9th IEEE International Symposium on Biomedical Imaging (ISBI). IEEE; 2012. p. 1735–1738.

\bibitem{bib20}
Bogovic JA, Hanslovsky P, Wong A, Saalfeld S.
\newblock {Robust registration of calcium images by learned contrast synthesis}.
\newblock . In: 2016 IEEE 13th international symposium on biomedical imaging (ISBI). IEEE; 2016. p. 1123–1126.

\bibitem{bib21}
Arganda-Carreras I, Kaynig V, Rueden C, Eliceiri KW, Schindelin J, Cardona A, et al. 
\newblock {Trainable Weka Segmentation: a machine learning tool for microscopy pixel classification}.
\newblock Bioinformatics. 2017;33(15):2424–2426.

\bibitem{bib22}
Roels J, Vernaillen F, Kremer A, Gon{\c{c}}calves A, Aelterman J, Luong HQ, et al.
\newblock {An interactive ImageJ plugin for semi-automated image denoising in electron microscopy}.
\newblock Nature communications. 2020;11(1):1–13.

\bibitem{bib23}
Sedzicki J, Tschon T, Low SH, Willemart K, Goldie KN, Letesson JJ, et al.
\newblock {3D correlative electron microscopy reveals continuity of Brucella-containing vacuoles with the endoplasmic reticulum}.
\newblock Journal of cell science. 2018;131(4).

\bibitem{bib24}
Paul-Gilloteaux P, Heiligenstein X, Belle M, Domart MC, Larijani B, Collinson L, et al.
\newblock {eC-CLEM: flexible multidimensional registration software for correlative microscopies}.
\newblock Nature methods. 2017;14(2):102.

\bibitem{bib25}
Velasco FR. 
\newblock {Thresholding using the ISODATA clustering algorithm}.
\newblock 1979.


\end{thebibliography}

\end{document}